\documentclass[a4paper]{article}

\usepackage{INTERSPEECH2019}
\usepackage{makecell}

\title{Learning Speaker Embedding with Momentum Contrast}
\name{Ke Ding, Xuanji He, Guanglu Wan}
\address{
  Meituan-Dianping Group}
\email{\{dingke02,\,hexuanji,\,wanguanglu\}@meituan.com}

\begin{document}

\maketitle
\begin{abstract}
  
Speaker verification can be formulated as a representation learning task, where speaker-discriminative embeddings are extracted from utterances of variable lengths.
Momentum Contrast (MoCo) is a recently proposed unsupervised representation learning framework, 
and has shown its effectiveness for learning good feature representation for downstream vision tasks.
In this work, we apply MoCo to learn speaker embedding from speech segments.
We explore MoCo for both unsupervised learning and pretraining settings.
In the unsupervised scenario, embedding is learned by MoCo from audio data without using any speaker specific information.
On a large scale dataset with $2,500$ speakers, MoCo can achieve EER $4.275\%$ trained unsupervisedly, and the EER can decrease further to $3.58\%$ if extra unlabelled data are used.
In the pretraining scenario, encoder trained by MoCo is used to initialize the downstream supervised training.
With finetuning on the MoCo trained model, the equal error rate (EER) reduces $13.7\%$ relative  ($1.44\%$ to $1.242\%$) compared to a carefully tuned baseline training from scratch.
Comparative study confirms the effectiveness of MoCo learning good speaker embedding.
\end{abstract}


\section{Introduction}

Speaker verification (SV) is the task of confirming the
claimed identity of a speaker given one's speech segments.
Typically, a fixed-dimensional embedding is extracted for both
enrollment and test speech, and compared to give out a same-speaker-or-not decision. 

I-vectors\,\cite{dehak2010ivector} are widely used as speaker embeddings.
In the standard i-vector system, a GMM is trained with all training data served as universal background model (UBM) to collect sufficient statistics (typically a super vector) from speech segments.
A low-rank project matrix, dubbed Total Variability Matrix, is trained with the sufficient statistics.
For a given segment, a super vector is extracted by UBM, and projected by the learned project matrix, resulting in a fixed-dimensional vector, i.e. i-vector.
The similarity between two i-vectors (usually the log likelihood ratio) can be computed by probabilistic linear discriminant analysis (PLDA)\,\cite{ioffe2006plda} .

In recent years, various neural network based methods are proposed to learning more discriminative embeddings.

X-vectors are proposed to replace i-vectors\,\cite{snyder2017xvector}. 
In the x-vector system, speaker embeddings are extracted by time-delay neural networks (TDNNs).
The TDNN is trained as a multi-speaker classifier using cross entropy loss and then the activation from some hidden layer is extracted as the embedding.
A temporal statistical pooling layer to applied to tackle segments of variable length.
After training, utterances are mapped directly to fixed-dimensional speaker embeddings, just as i-vector systems do.
Working with a PLDA backend, the x-vectors can outperform i-vectors for short speech segments and are competitive on long duration test conditions\,\cite{snyder2017xvector,snyder2018xvector}.

The embeddings learned by cross entropy loss is separable by design for the close-set classification task, but not necessarily discriminative enough, which is key to generalize to identify unseen speaker segments.
To tackle this issue, various training criterion to enhance the discriminative power of the x-vector embeddings.

Contrastive learning, i.e. contrastive loss\,\cite{chen2011siamese}, triplet loss\,\cite{zhang2018triplet}, is introduced for optimizing the distances between embeddings directly.
Losses of constrastive style minimize the distance between an anchor and a positive sample while maximizes the distance between the anchor and a negative one, thus encouraging embeddings with compact within-speaker variations and separable between-speaker differences.


However, optimizing contrastive loss can be challenging\,\cite{zhang2018triplet,ghosh2018centerloss},
and selecting training pairs or triplets suffers from combinatorial data expansion and negative samples mining is necessary for effective and stable training.
Margin based training criterion is proposed to learn embeddings with compact inter-class variation and bypass the training problems of constrastive learning.

Center loss\,\cite{li2018centerloss} is used to work with cross entropy loss (to avoid embedding collapsing).
Center loss penalises the distance between the embeddings and their corresponding class centers in the Euclidean space to achieve intra-class compactness.
By introducing margins between classes into conventional cross entropy loss, angular softmax (A-softmax) is reported being able to learn more discriminative embeddings than cross entropy loss and triplet loss\,\cite{li2018asoftmax}.
More recently, Additive Angular Margin (AAM) loss is proposed for extracting highly discriminative features for face recognition\,\cite{deng2019arcface}.
AAM is successfully applied to speaker verification task\,\cite{xiang2019aam} and achieves state of the art performance in the VoxCeleb challenge\,\cite{zeinali2019but}.

 
Recently, a novel unsupervised learning framework, Momentum Contrast (MoCo) \,\cite{he2019moco} is proposed.
MoCo is an extension of instance discrimination, and can learn representation using contrastive learning criterion.
In several vision tasks, MoCo can outperform its supervised pretraining counterpart, thus largely closing the gap between unsupervised and supervised representation.

In this work, we apply MoCo to the speaker verification task.
Observe that MoCo trains an encoder directly, which is exactly the conventional neural network based methods (e.g. x-vector system) do.
What's more, MoCo encourages discriminative representation (via contrastive learning), which is key for open set verification.
We explore MoCo in the ways: 1) using MoCo trained encoder directly, 2) using MoCo as a pretraining method to relieve downstream training of interest.

The remainder of this paper is organized as follows.
Section\,\ref{sec:proposal} describes our proposal for using  MoCo for learning speaker embedding.
Specifically, a SpecAugment\,\cite{park2019specaugment} based data augmentation is introduced for parallel speech segments generation.
Section\,\ref{sec:exp} shows the experiments conducted on a large scale speaker dataset.
Results on the public available Voxceleb dataset are also presented.
We conclude our work in Section\,\ref{sec:conclusion}.

\section{Proposed Method}
\label{sec:proposal}

\subsection{Momentum Contrast Learning}
\label{ssec:moco}

\begin{figure}[t]
    \centering
    \includegraphics[width=0.7 \linewidth]{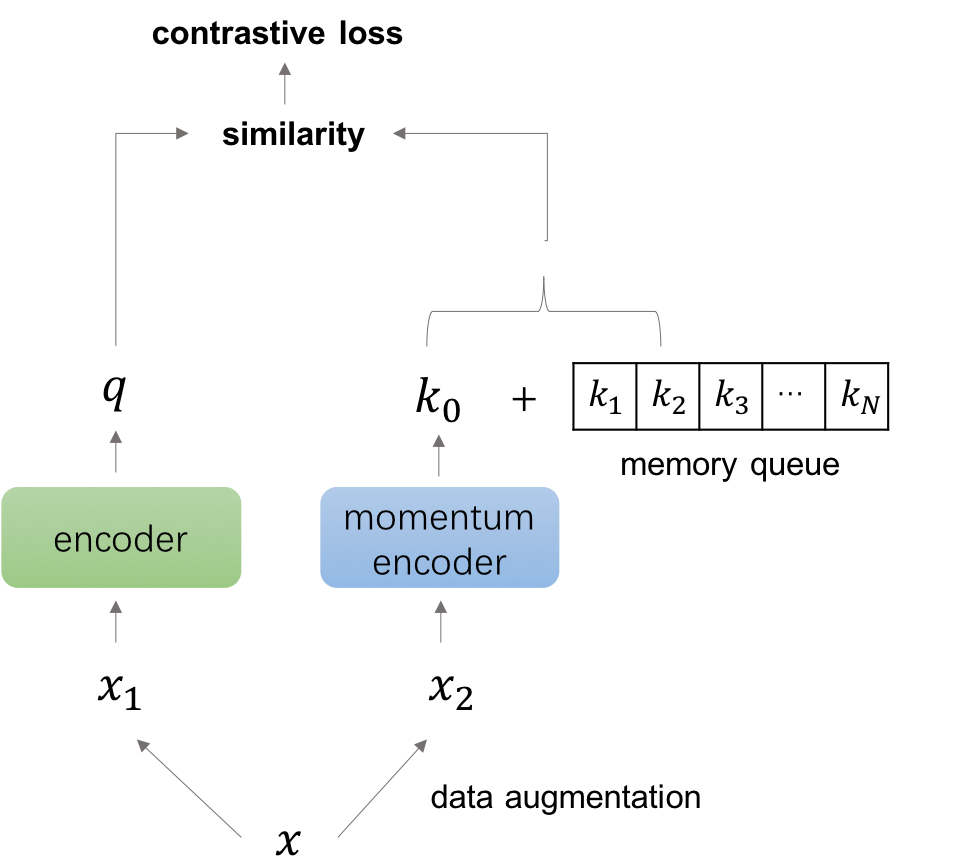}
    \caption{Training framework of MoCo}
    \label{fig:moco}
\end{figure}

Momentum Contrast (MoCo)\,\cite{he2019moco} is a general mechanism for unsupeverised learning representation using contrastive loss.
In MoCo, a dynamic dictionary is built with a queue and a moving-averaged encoder. 
This enables building a large and consistent dictionary on-the-fly that facilitates contrastive unsupervised learning.
The representations learned by MoCo is reported to transfer well to downstream visual tasks. 

The training framework of MoCo is depicted by Fig.\,\ref{fig:moco}.
For each training sample, two corrupted versions are generated by some augmentation strategy.
After processing one mini-batch, the encoder's parameters are updated by some optimizer, say SGD.
The encoded representations of the current mini-batch ($k_0$ in Fig.\,\ref{fig:moco}) are enqueued, and the oldest ones are dequeued to keep the queue size consistent.
Before processing the next mini-batch, the momentum encoder is update with the encoder with as momentum coefficient (typically close to 1).

As neural network based speaker verification systems learn embedding from utterances, MoCo can be applied to the training process in a natural way.
To be specific, the encoder is as same as the network used for conventional xvector training (see Sec.\,\ref{sec:exp}).
We just initialize the momentum encoder with the encoder and initialize memory queue randomly.
For other details and import tricks for training (i.e. constrastive loss, ShuffleBN), we refer the reader to\,\cite{he2019moco}.

\subsection{Data Augmentation}
\begin{figure}[t]
    \centering
    \includegraphics[width=0.7\linewidth]{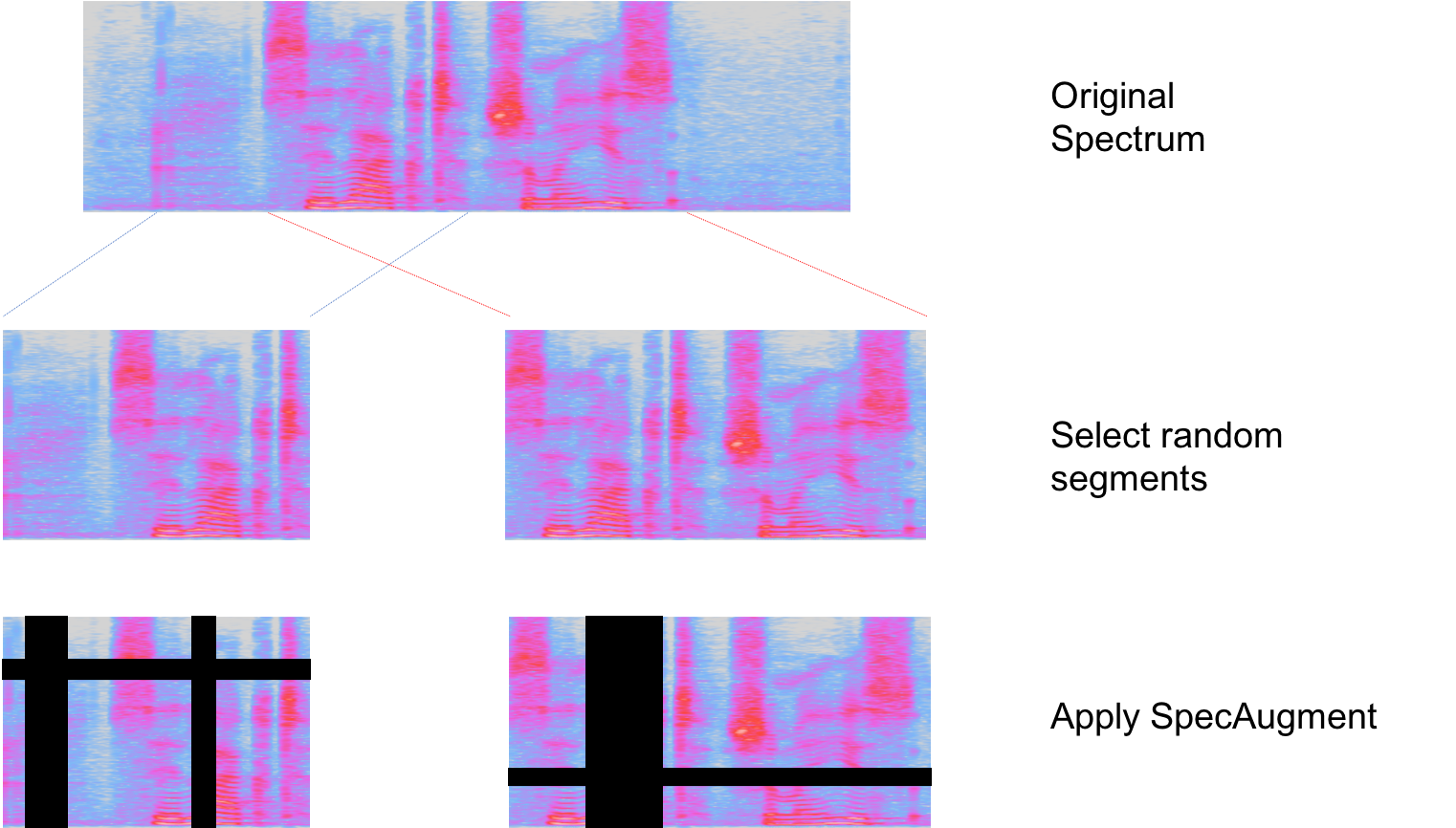}
    \caption{Generating two augmented version of segments from one utterance}
    \label{fig:agument}
\end{figure}

What's specific to our task under study (speaker verification) is how to generate parallel corrupted version of the same utterance.
The method we propose is depicted by Fig\,\ref{fig:agument}.
First, we randomly selected two segments from the target utterance, which is the common practice in x-vector training.
Second, we apply SpecAugment\,\cite{park2019specaugment} to the segments, resulting in two different version of the same utterance with various length and spectrum distortion.
Via the proposed process, parallel corrupted segments can be generated with both temporal and spectrum variability.

\subsection{MoCo as embedding extractor}
 
MoCo trains the encoder as an instance discrimination task, treating the parallel corrupted version as positive sample and all in the memory queue as negative samples.
So it's natural to expect the embeddings extracted by a well trained encoder show discrimination between different speakers.
Reminiscent of the conventional i-vector extractor, we can use the learned encoder with some backend (e.g. PLDA, cosine) for verification task directly.
As MoCo need no speaker information, we also explore if extra unlabelled data can lead to more robust embedding for verification.

\subsection{MoCo as Pretraining}

Though unsupervised learning may not be optimal for the task of interest, features pretrained by unsupervised learning might be transferred to downstream tasks by fine-tuning.
In the previous work, cross entropy training is used for pretraining\,\cite{zeinali2019but}.
In this work, we explore if models unsupervised pretrained by MoCo is helpful for the downstream supervised learning.

\section{Experiments}
\label{sec:exp}

\subsection{Datasets}
\label{ssec:datasets}
We train and evaluate our models on two datasets, Dataset A with speaker labels and Dataset B without any speaker information.

\textbf{Dataset A} consists of utterances from $2,943$ speakers, each with $1,000$ utterances.
All speech are recorded with mobile phones (iOS or Android systems) with a sample rate $16$\,kHz.
The duration of utterances  are distributed between $2.5$s and $6.5$s, with a mode duration 3.5s.
We split the dataset into training and test set.
The training set consists of $2,500$ speakers.
For each of the remaining $443$ speakers, we randomly selected three utterances for enrollments, and $50$ utterance for evaluation, resulting in about $9.8$ million trials.
Other remaining utterances of the test speakers are not used in the experiments.

\textbf{Dataset B} consists of mobile phone recorded speech with sample rate $16$\,kHz, without any speaker specific information. 
There are $7.6$ million utterances in total, or about $9$ thousand hours before silences removing.

The information of the two datasets are summarized in Tab.\,\ref{tab:datasets}.

\begin{table}[]
    \centering
    \begin{tabular}{c|c|c}
    \toprule
       dataset  & training  &  test  \\
\midrule
      A  & \thead{$2,500$ speakers \\ $1,000$ utterances} & \thead{$443$ speakers\\ ($3$ for enrollment, \\$50$ for evaluation)} \\
      \hline
      B & \thead{ $\sim7.6$ million utterances \\  $\sim9$k hours} & - \\
      \bottomrule
    \end{tabular}
    \caption{Overview of the training and evaluation data}
    \label{tab:datasets}
\end{table}

\subsection{Baseline systems}
\label{ssec:baseline}

\subsubsection{I-vector system}

The i-vector system is based on the standard kaldi recipe \cite{povey2011kaldi} (egs/voxceleb/v1).
The features are 30-dimensional Mel-Frequency Cepstral Coefficients (MFCCs), with a frame shift of 10ms and a window width of 25ms. 
Delta and acceleration are appended to create 90 dimension feature vectors. 
Mean normalization is then applied over a sliding window of up to 3 seconds. 
To filter out non-speech frames, kaldi's energy-based voice activity detector (VAD) is employed. 
(We also tried a neural network based VAD, but observe no improvement.)
The UBM is a 2048 component full-covariance GMM. 
The system uses a 400 dimension i-vector extractor. 
I-vectors are centered, dimensionality reduced to 200 using LDA.
Length normalization\,\cite{garcia2011ivector} is applied before PLDA scoring\,\cite{ioffe2006plda}.

\subsubsection{X-vector system}
\label{ssec:xvector}

\begin{table}[]
    \centering
    \begin{tabular}{c  c  c  c}
    \toprule
         \textbf{Layer} & \textbf{Layer Context}  & \textbf{Input $\times$ Output}  \\
         \midrule
         frame1 & [t-2, t+2] & (5 $\times$ d) $\times$ 512 \\
         frame2 & \{t-2, t, t+2\}  & (3 $\times$ 512) $\times$ 512 \\
         frame3 & \{t-3, t, t+3\}  & (3 $\times$ 512) $\times$ 512 \\
         frame4 & \{t\} & 512 $\times$ 512 \\
         frame5 & \{t\} & 512 $\times$ 1500 \\
         stats pooling & [0, T) & 1500T $\times$ 3000 \\
         embed\_a & \{0\} & 3000 $\times$ 512 \\
         embed\_b & \{0\}  & 512 $\times$ 512 \\
         softmax & \{0\} & 512 $\times$ D \\
         \bottomrule
    \end{tabular}
    \caption{X-vector network architecture used in this paper, where d, T, and D denote the dimensionality of input feature, the number of utterance frames, and the number of speakers, respectively. Embeddings are extracted at layer embed\_b (before nonlinearity).}
    \label{tab:xvector}
\end{table}
Our x-vector is based kaldi's voxceleb recipe.
The TDNN structure used is as Tab.\,\ref{tab:xvector}.
We train the x-vector network using pytorch\,\cite{paszke2019pytorch}, over four V100 GPUs.
Synchronous Stochastic Gradient Descent (SGD) with weight decay $=1e-5$ and momentum $=0.9$ is used.
The total batch size is $1024$ ($256$ for each GPU).
The learning rate  decays from $1e-4$ to $1e-5$ exponentially.
A dropout layer\,\cite{srivastava2014dropout} with $p=0.5$ is applied after before embed\_b.
A max gradient norm of $2$ is applied to stabilize training.

After silence removing, we filter out utterances shorter than $250$ frames and make sure each speaker has at least $30$ utterances,
ending up with $2,453$ speaker, about $1184$ hours.
We train $30$ times of the total training frames, with each epoch processing $72$ million frames (i.e. $200$ hours).
We random select 3 utterances from each speaker as holdout validation set.

Like the ivector-system, a PLDA is used as backend for scoring.
The LDA reduction dimension is set to $150$.

\subsubsection{X-vector with AAM loss}

Additive Angular Margin (AAM) loss is reported more effective than the conventional cross entropy loss for both face\,\cite{deng2019arcface} and speaker verification tasks\,\cite{zeinali2019but}.

AAM loss is defined as Eq.\,\ref{eq:aam}:
\begin{equation}
\label{eq:aam}
 L = -\frac{1}{N}\sum_{i=1}^{N}\log \frac{e^{s(\cos(\theta_{y_i}+m))}}{e^{s(\cos(\theta_{y_i}+m))} + \sum_{j=1,j\neq y_i}^{D} e^{s\cos\theta_{j}}}
\end{equation}
where $\cos(\theta)$ is 
the dot product between the embed\_b and the last fully connected layer (softmax layer) (after L2 normalization), additive angular $m$ and scale factor $s$ are hyper parameters. 

In our experiments, layers after embed\_b are dropped and an AAM loss is applied instead of cross entropy loss.
We use $s = 32$, $m = 0.3$ for all experiments.
The learning rate  decays from $1e-5$ to $1e-6$ exponentially.
No dropout is used, as we found it degrades the performance.
A max gradient norm of $6$ is applied to stabilize training.
All other hyperparameters stay the same as the cross entropy training.

We try two backend, 1) cosine distance and 2) PLDA as the aforementioned the x-vector system.

The EERs and minDCFs of the three baseline systems are listed in Tab\,\ref{tab:baseline}. 
It can be found that AAM can achieve better performances than cross entropy based training, and this observation is in line with findings by\,\cite{xiang2019aam,zeinali2019but}.
Direct cosine distance can achieve decent performance, and a separate PLDA backend contribute no improvement (though no significant performance degradation w.r.t EER, is surpassed by cosine on minDCF criterion by large margins). 
Thus, we use AAM-loss trained network with cosine backend for future comparison.

\begin{table*}[]
    \centering
    \begin{tabular}{cccccc}
    \toprule
       method & ivector (cosine)  & ivector (PLDA) &  xvector-CE (PLDA) & xvector-AAM (PLDA) & xvector-AAM (cosine) \\
    \midrule
      EER ($\%$)  & 5.178 & 1.661 & 1.562 & 1.589 & \textbf{1.44} \\
      \hline
      minDCF$_{0.01}$ & 0.397 & 0.191 & 0.217 & 0.24 & \textbf{0.165} \\
      \hline
      minDCF$_{0.001}$ & 0.622 & 0.409 & 0.487 & 0.572 & \textbf{0.346} \\
      \bottomrule
    \end{tabular}
    \caption{Performance for the baseline systems.}
    \label{tab:baseline}
\end{table*}

\subsection{MoCo as extractor}
\label{ssec:extractor}
\begin{table}[t]
    \centering
    \begin{tabular}{ccc}
    \toprule
        method & cosine & LDA-PLDA  \\
        \midrule
        ivector & 5.178 & \textbf{1.661} \\
        MoCo & 4.275 & 2.655 \\
        MoCo (+ Dataset B) & \textbf{3.58} & 2.366 \\
        \bottomrule
    \end{tabular}
    \caption{EERs ($\%$) of encoder learned by MoCo with different backends.}
    \label{tab:unsuper}
\end{table}

The network used for x-vector system (c.f. Sec.\,\ref{ssec:xvector}) is used as the MoCo encoder.
We use a queue size $10,000$, and $\beta=0.99$ $\tau=0.07$.
For SpecAugment, time warp window $10$, max time mask width $20$, max frequency mask width $10$.
The learning rate decays exponentially from $1e-4$ to $1e-5$.

After training, the encoder is used for xvector extraction.
As in the xvector-AAM case, we try both cosine and PLDA backends.
The results are listed in Tab.\,\ref{tab:unsuper}.

We compare MoCo with i-vector system, as both MoCo and i-vevtor's training process is unsupervised.
As can be seen, with cosine backend (complete unsupervised), MoCo outperforms i-vector.
It is not surprising, as MoCo optimizes dot product (cosine) as the training criterion.
Interestingly, helped by a strong backend (PLDA), i-vector can achieve an EER $1.661\%$, significantly better than MoCo ($2.655\%$).

It is not clear whether some backend other than PLDA used here can improve the performance for MoCo.
We leave this topic for future work.

To explore if extra data is helpful for MoCo learning, we train MoCo use extra unlabelled data from Dataset B.
We use a larger queue size ($20,000$), and other hyperpamaters stay the same.
The performance of MoCo encoder can improve further, from $4.275\%$ to $2.655\%$ for the cosine backend, and $3.58\%$ to $2.366\%$ for the PLDA backend.

\subsection{MoCo as pretraining}
\label{ssec:pretraining}

\begin{table}[]
    \centering
    \begin{tabular}{cccc}
    \toprule
       method  & EER ($\%$)  &  minDCF$_{0.01}$ & minDCF$_{0.001}$ \\
    \midrule
      xvector-AAM & 1.44 & 0.165 & 0.346 \\
      \hline
      \begin{tabular}{@{}c@{}}
      xvector-AAM \\ (CE)
      \end{tabular} & 1.481 & 0.169 & 0.357 \\
      \hline
     \begin{tabular}{@{}c@{}}
      xvector-AAM\\ (MoCo)
      \end{tabular} & 1.242 & 0.157 & 0.334 \\
     \hline
     \begin{tabular}{@{}c@{}}
      xvector-AAM\\ (MoCo-extra)
      \end{tabular} & \textbf{1.223} & \textbf{0.146} & \textbf{0.312} \\

      \bottomrule
      
    \end{tabular}
    \caption{Results of AAM trained models using different pretraining strategy. CE, Moco, MoCo-extra stands for, respectively,cross entropy pretraining, pretraining with Dataset A, and with both Dataset A and Dataset B. Cosine backend is applied for all models}
    \label{tab:pretraining}
\end{table}

In this section, we study if the encoder trained with MoCo is helpful for downstream supervised learning.
We initialize x-vector with the encoder pretrained by MoCo in Sec.\,\ref{ssec:extractor}.
As comparison, we also conduct experiment with conventional pretraining with cross entropy training (as used in\,\cite{zeinali2019but}).
As shown in Tab\,\ref{tab:pretraining}, model pretrained by MoCo can great help the training.
The EER improves from $1.44\%$ to $1.242\%$, a $13.7\%$ relative improvement.
At the same time, we find that no improvement is observed with cross entropy pretraining.
Cross entropy pretraining does not transfer well to the downstream AAM training.
In fact it is harmful for AAM training in our case.
One possible explanation is that both MoCo and AAM encourage the representation to hyper-sphere subspace (via softmax over dot product),
while cross entropy learns a difference subspace.

\begin{figure}
    \centering
    \includegraphics[width=0.58\linewidth]{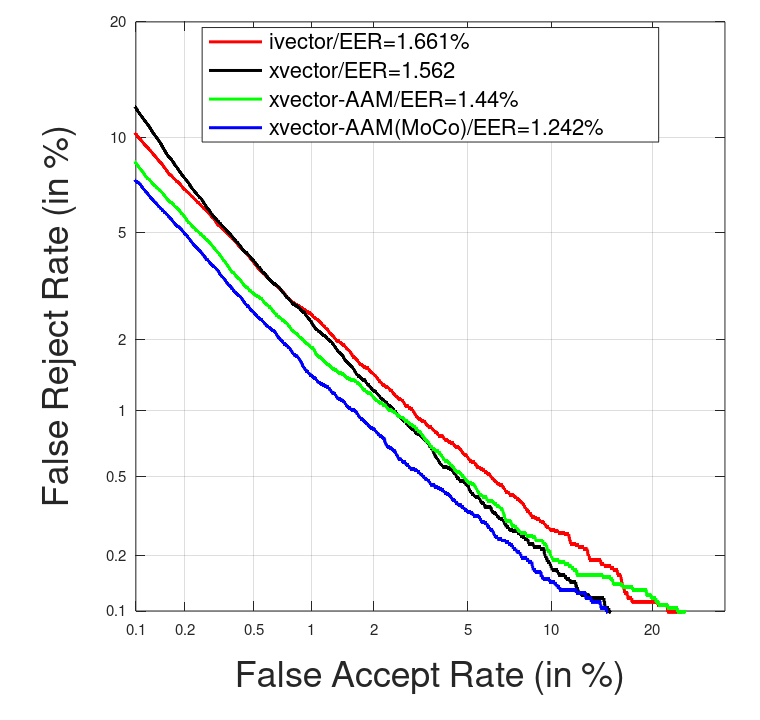}

    \caption{The DET curves of Ivector and Xector models trained with different losses and strategies trained on Dataset A.}
    \label{fig:det}
\end{figure}

The DET curves of the systems under study are illustrated in Fig.\,\ref{fig:det}.
Moco pretraining improves the both EER and minDCF criteria; in the meantime, it achieves better performance under different false accept and false reject tradeoffs.

\subsection{Results on Voxceleb}

To check if the observations on our in-house data generalize, we conduct experiments on the public available dataset Voceleb. 
All data from Voxceleb2\,\cite{chung2018voxceleb} and the training part to Voxceleb1\,\cite{nagrani2017voxceleb} are used for training.
After converting the audios from amm to wav format, it ends up with $7,323$ speaker and $1,276,888$ utterances in total.
Voxceleb1 test is used as the test, which has $4,874$ utterance from $40$ speakers.
The official test protocol is used ($37,720$ trials in total).

The same network architecture and training configuration as previous experiments are used, and no further hyper parameter tuning is conducted for this task.
According to Tab\,\ref{tab:voxceleb}, similar tendency with the previous experiments is observed.
MoCo pretraining does help AAM training, while CE pretraining doesn't
Besides, the results confirms that embeddings learnned by AAM works better with cosine backend. 

\begin{table}[]
    \centering
    \begin{tabular}{cccc}
    \toprule
       method  & EER ($\%$)  &  minDCF$_{0.01}$ & minDCF$_{0.001}$ \\
    \midrule
      \begin{tabular}{@{}c@{}}
      ivector \\ (PLDA)
      \end{tabular} & 5.467 & 0.4859 & 0.6213 \\
      \hline
      \begin{tabular}{@{}c@{}}
      xvector-CE \\ (PLDA)
      \end{tabular} & 3.356 & 0.3591 & 0.5890 \\
      \hline
      \begin{tabular}{@{}c@{}} xvector-AAM \\ (cosine) \end{tabular} & 2.497 & 0.2634 & 0.3888 \\
      \hline
      \begin{tabular}{@{}c@{}}
      xvector-AAM \\ (PLDA)
      \end{tabular} & 2.752 & 0.3717 & 0.4971 \\
      \hline
      \begin{tabular}{@{}c@{}} xvector-AAM \\(CE pretraing)\\(cosine) \end{tabular} & 2.572 & 0.2535 &  0.3799 \\
      \hline
      \begin{tabular}{@{}c@{}} xvector-AAM \\(MoCo pretraing)\\(cosine) \end{tabular}  & \textbf{2.402} & \textbf{0.2476} & \textbf{0.3506} \\
      \bottomrule
      
    \end{tabular}
    \caption{Results on VoxCeleb1 trained on VoxCeleb1 dev set and all VoxCeleb2 data.}
    \label{tab:voxceleb}
\end{table}

\section{Conclusions}
\label{sec:conclusion}

In this paper, we explore the effectiveness of MoCo for learning speaker embedding.
To apply MoCo with speech data, we propose a parallel distorted data generating strategy based on SpecAugment.
The experiments show that MoCo learns good speaker discriminative embedding, and can be used as an effective pretraining method for the downstream supervised training.
On a large-scale dataset, we build a strong baseline system with AAM, which can significantly outperform the conventional cross entropy based x-vector system.
Compared to the baseline system trained from scratch, MoCo pretraining can achieve a $13.7\%$ relative EER improvement while the conventional cross entropy pretraining gains no improvement.
Thanks to MoCo's unsupervised nature, extra unlabelled data could be used to improve the performance further.

\bibliographystyle{IEEEtran}

\bibliography{mybib}

\end{document}